# A WORKSPACE BASED CLASSIFICATION OF 3R ORTHOGONAL MANIPULATORS

Philippe Wenger, Maher Baili and Damien Chablat

*Institut de Recherche en Communications et Cybernétique de Nantes, UMR CNRS 6597*
*1 rue de la Noë, BP 92101, 44321 Nantes Cedex 3, France*

Philippe.Wenger@irccyn.ec-nantes.fr

**Abstract**   A classification of a family of 3-revolute (3R) positioning manipulators is established. This classification is based on the topology of their workspace. The workspace is characterized in a half-cross section by the singular curves of the manipulator. The workspace topology is defined by the number of cusps and nodes that appear on these singular curves. The design parameters space is shown to be partitioned into nine subspaces of distinct workspace topologies. Each separating surface is given as an explicit expression in the DH-parameters.

**Keywords:**   Classification, Workspace, Singularity, Cusp, node, orthogonal manipulator

## 1.     Introduction

This paper focuses on positioning 3R manipulators with orthogonal joint axes (orthogonal manipulators). Orthogonal manipulators may have different global kinematic properties according to their link lengths and joint offsets. Unlike usual industrial manipulators, orthogonal manipulators may be cuspidal, that is, they can change their posture without meeting a singularity (Parenti and Innocenti, 1988, Burdick, 1988). This property was unknown before 1988 (Borrel and Liégeois, 1988). Several years later, some conditions for a manipulator to be noncuspidal were provided, which include simplifying geometric conditions like parallel and intersecting joint axes (Burdick, 1995) and also nonintuitive conditions (Wenger, 1997). A general necessary and sufficient condition for a 3-DOF manipulator to be cuspidal was established in (El Omri and Wenger, 1995), namely, the existence of at least one point in the workspace where the inverse kinematics admits three equal solutions. The word "cuspidal manipulator" was defined in accordance to this condition because a point with three equal IKS forms a cusp in a cross section of the workspace (Arnold, 1981). The categorization of all generic 3R manipulators was established in (Wenger, 1998) based on the homotopy class of the singular curves in the joint space. Wenger, 1999 proposed a procedure to take into account the cuspidality property in the design process of new manipulators. More recently, Corvez and Rouillier, 2002 applied efficient algebraic tools to the classification of 3R orthogonal manipulators with no offset on their last joint. Five surfaces were found to divide the parameters space into 105 cells with the same number of cusps in the workspace. The equations of these five surfaces were derived as polynomials in the DH-parameters using Groebner Bases. A kinematic interpretation of this theoretical work showed that, in fact, only five different domains exist : two domains of noncuspidal manipulators, one domain where manipulators have two cusps and two domains where they have four cusps (Baili et al, 2003). However, the authors did not provide the equations of the true separating surfaces in the parameters space. On the other hand, they did not take into account the occurrence of nodes, which play an important role for analyzing the number of IKS in the workspace.

The purpose of this work is to classify a family of 3R positioning manipulators according to the topology of their workspace, which is defined by the number of cusps and nodes that appear on the singular curves. The design parameters space is shown to be divided into nine domains of distinct workspace topologies. In each domain, the distribution of the number of IKS is the same. This study is of interest for the design of new manipulators.

The rest of this article is organized as follows. Next section presents the manipulators under study and recalls some preliminary results. The classification is established in section 3. Section 4 synthesizes the results and section 5 concludes this paper.

## 2. Preliminaries

### 2.1  Manipulators under study

The manipulators studied in this paper are orthogonal with their last joint offset equal to zero. The remaining lengths parameters are referred to as $d_2$, $d_3$, $d_4$, and $r_2$ while the angle parameters $\alpha_2$ and $\alpha_3$ are set to $-90°$ and $90°$, respectively. The three joint variables are referred to as $\theta_1$, $\theta_2$ and $\theta_3$, respectively. They will be assumed unlimited in this study. Figure 1 shows the kinematic architecture of the manipulators under study in the zero configuration. The position of the end-tip (or wrist center) is defined by the three Cartesian coordinates $x$, $y$ and $z$ of the operation point $P$ with respect to a reference frame (O, **x**, **y**, **z**) attached to the manipulator base as shown in Fig. 1.

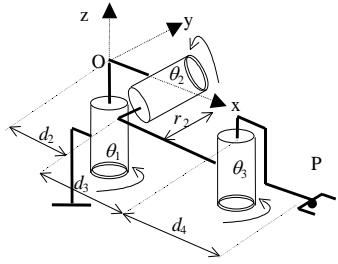
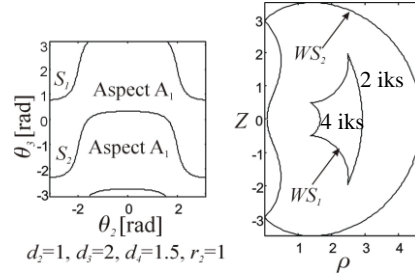

*Figure 1*. Orthogonal manipulators under study.

*Figure 2*. Singularity curves in joint space (left) and workspace (right)

## 2.2. Singularities and aspects

The determinant of the Jacobian matrix of the orthogonal manipulators under study is $\det(\mathbf{J}) = (d_3 + c_3 d_4)(s_3 d_2 + c_2(s_3 d_3 - c_3 r_2))$, where $c_i=\cos(\theta_i)$ and $s_i=\sin(\theta_i)$. A singularity occurs when $\det(\mathbf{J})=0$. The contour plot of $\det(\mathbf{J})=0$ forms a set of curves in $-\pi \leq \theta_2 < \pi, -\pi \leq \theta_3 < \pi$.

If $d_3 > d_4$, the first factor of $\det(\mathbf{J})$ cannot vanish and the singularities form two distinct curves $S_1$ and $S_2$ in the joint space (El Omri, 1996), which divide the joint space into two singularity-free open sets $A_1$ and $A_2$ called *aspects* (Borrel and Liégeois, 1988). The singularities can also be displayed in the Cartesian space (Kholi and Hsu, 1987, Ceccarelli, 1996). Thanks to their symmetry about the first joint axis, a 2-dimensional representation in a half cross-section of the workspace is sufficient. The singularities form two disjoint sets of curves in the workspace. These two sets define the internal boundary $WS_1$ and the external boundary $WS_2$, respectively, with $WS_1=f(S_1)$ and $WS_2=f(S_2)$. Figure 2 shows the singularity curves when $d_2=1$, $d_3=2$, $d_4=1.5$ and $r_2=1$. For this manipulator, the internal boundary $WS_1$ has four cusp points. It divides the workspace into one region with two IKS (the outer region) and one region with four IKS (the inner region).

If $d_3 \leq d_4$, the operation point can meet the second joint axis whenever $\theta_3 = \pm\arccos(-d_3/d_4)$ and two horizontal lines appear in the joint space $-\pi \leq \theta_2 < \pi, -\pi \leq \theta_3 < \pi$, which may intersect $S_1$ and $S_2$ depending on $d_2$, $d_3$, $d_4$ and $r_2$ (El Omri, 1996). The number of aspects depends on these intersections. Note that if $d_3 < d_4$, no additional curve appears in the workspace cross-section but only two points. This is because, since the operation point meets the second joint axis when $\theta_3=\pm\arccos(-d_3/d_4)$, the location of the operation point does not change when $\theta_2$ is rotated.

## 3. Workspaces classification

## 3.1 Classification criteria

The classification is conducted on the basis of the topology of the singular curves in the workspace, which we characterize by (i) the number of cusps and (ii) the number of nodes. A cusp (resp. a node) is associated with one point with three equal IKS (resp. with two pairs of equal IKS). These singular points are interesting features for characterizing the workspace shape and the distribution of the number of IKS in the workspace.

## 3.2 Number of cusps

For now on and without loss of generality, $d_2$ is set to 1. Thus, only three parameters $d_3$, $d_4$ and $r_2$ need to be handled. Baili et al, 2003 showed that one or more surfaces among the five ones found by Corvez and Rouillier, 2002, are not relevant. However, they did not try to find which surfaces are really separating. To derive the equations of the separating surfaces, we investigate the transitions between the five domains. First, let us recall the five different manipulator types associated with the five domains found by Corvez and Rouillier, 2002. The first type is a binary manipulator (i.e. it has only two IKS) with no cusp and a hole (Fig. 3). The remaining four types are quaternary manipulators (i.e. with four IKS). The second type is a manipulator with four cusps on the internal boundary.

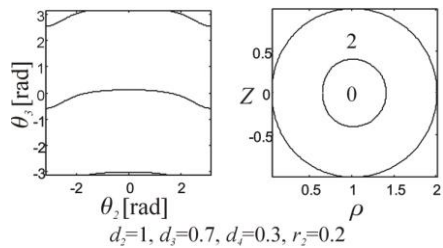
$d_2=1$, $d_3=0.7$, $d_4=0.3$, $r_2=0.2$
*Figure* 3. Manipulator of type 1.

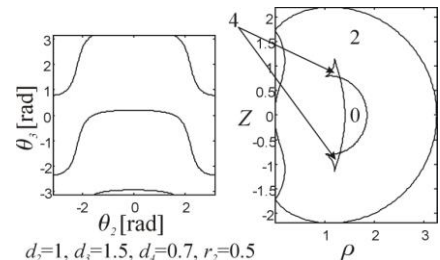
$d_2=1$, $d_3=1.5$, $d_4=0.7$, $r_2=0.5$
*Figure* 4. Manipulator of type 2.

There are three instances of type 2 according to the number of nodes, as will be shown in section 3.3. The first one is shown in Fig. 4 with two nodes. The second one was shown in Fig. 2 with no node. The last one is such that $d_3 \leq d_4$ and will be shown in Fig. 10.

Transition between type 1 and type 2 is a manipulator having a pair of points with four equal IKS, where two nodes and one cusp coincide. Proof of this result can be obtained as a straightforward consequence of transitions in quartics root patterns (Baili, 2003). Deriving the condition for the inverse kinematic polynomial to have four equal roots yields the

equation of the separating surface (Baili, 2003)

$$d_4 = \sqrt{\frac{1}{2}\left(d_3^2 + r_2^2 - \frac{(d_3^2 + r_2^2)^2 - d_3^2 + r_2^2}{AB}\right)} \quad (1)$$

where $A = \sqrt{(d_3+1)^2 + r_2^2}$ and $B = \sqrt{(d_3-1)^2 + r_2^2}$

The third type is a manipulator with only two cusps on the internal boundary, which looks like a fish with one tail (Fig. 5).

As shown by Baili, 2003, transition between type 2 and type 3 is characterized by a manipulator for which the singular line given by $\theta_3 = -\arccos(-d_3/d_4)$ is tangent to the singularity curve $S_1$. Expressing this condition yields the equation of the separating surface

$$d_4 = \frac{d_3}{1+d_3} \cdot A \quad (2)$$

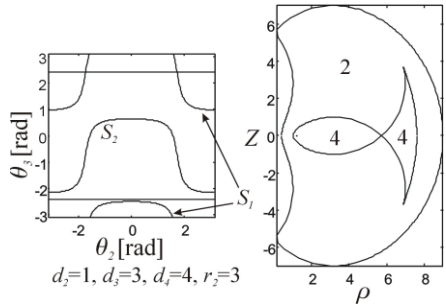
$d_2=1, d_3=3, d_4=4, r_2=3$

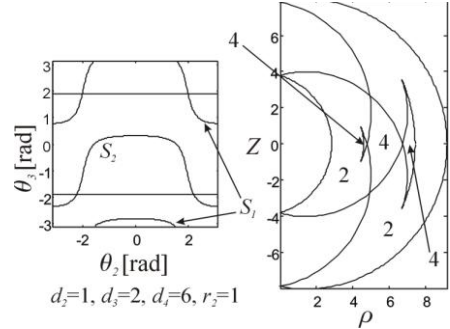
$d_2=1, d_3=2, d_4=6, r_2=1$

*Figure* 5. Manipulator of type 3.     *Figure* 6. Manipulator of type 4.

The fourth type is a manipulator with four cusps. Unlike type 2, the cusps are not located on the same boundary (Fig. 6).

Transition between type 3 and type 4 is characterized by a manipulator for which the singular line given by $\theta_3 = -\arccos(-d_3/d_4)$ is tangent to the singularity curve $S_2$ (Baili, 2003). Expressing this condition yields the equation of the separating surface

$$d_4 = \frac{d_3}{d_3-1} \cdot B \quad \text{and} \quad d_3 > 1 \quad (3)$$

Last type is a manipulator with no cusp (Fig. 7).

Unlike type 1, the internal boundary does not bound a hole but a region with 4 IKS. The two isolated singular points inside the inner region are associated with the two singularity lines. Transition between type 4 and type 5 is characterized by a manipulator for which the singular line given by $\theta_3 = +\arccos(-d_3/d_4)$ is tangent to the singularity curve $S_1$ (Baili, 2003).

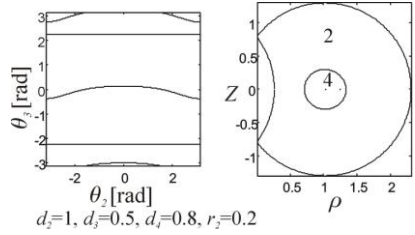

*Figure* 7. Manipulator of type 5.

Expressing this condition yields the equation of the separating surface

$$d_4 = \frac{d_3}{1-d_3} \cdot B \quad \text{and} \quad d_3 < 1 \tag{4}$$

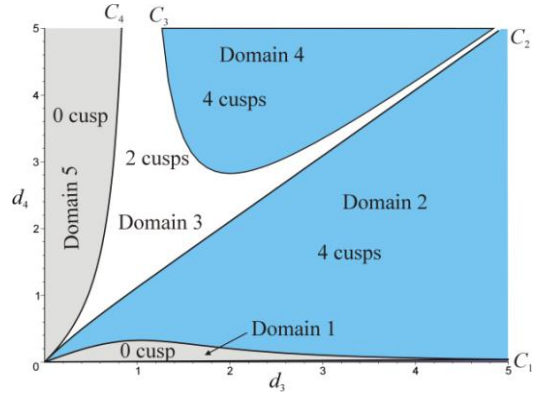

*Figure* 8. Plots of the four separating surfaces in a section ($d_3$, $d_4$) of the parameter space for $r_2=1$.

We have provided the equations of four surfaces that divide the parameters space into five domains where the number of cusps is constant. Figure 8 shows the plots of these surfaces in a section ($d_3$, $d_4$) of the parameter space for $r_2=1$. Domains 1, 2, 3, 4 and 5 are associated with manipulators of type 1, 2, 3, 4 and 5, respectively. $C_1$, $C_2$, $C_3$ and $C_4$ are the right hand side of (1), (2), (3) and (4), respectively. It is interesting to see the correspondence between the equations found with pure algebraic reasoning in (Corvez and Rouillier, 2002) and those provided in this paper. The five equations found by Corvez and Rouillier are:

$$-d_3 + d_4 r_2^2 + d_4 = 0 \tag{5}$$

$$d_3^2 - d_4^2 + r_2^2 = 0 \tag{6}$$

$$\begin{aligned} d_4^2 d_3^6 - d_4^4 d_3^4 + 3d_4^2 d_3^4 r_2^2 - 2d_4^2 d_3^4 + 2d_4^4 d_3^2 - 2d_4^4 d_3^2 r_2^2 + d_4^2 d_3^2 + 3d_4^2 d_3^2 r_2^4 \\ -d_3^2 r_2^2 - 2d_4^4 r_2^2 - d_4^4 r_2^4 - d_4^4 + d_4^2 r_2^6 + d_4^2 r_2^2 + 2d_4^2 r_2^4 = 0 \end{aligned} \tag{7}$$

$$d_3^2 r_2^2 + d_3^2 - 2d_3^3 + d_3^4 - d_4^2 + 2d_3 d_4^2 - d_3^2 d_4^2 = 0 \tag{8}$$

$$d_3^2 r_2^2 + d_3^2 + 2d_3^3 + d_3^4 - d_4^2 - 2d_3 d_4^2 - d_3^2 d_4^2 = 0 \tag{9}$$

Equation (7) is a second-degree polynomial in $d_4^2$. Solving this quadratics for $d_4$ shows that (7) has the following two branches:

$$d_4 = \sqrt{\frac{1}{2}\left(d_3^2 + r_2^2 - \frac{(d_3^2 + r_2^2)^2 - d_3^2 + r_2^2}{AB}\right)} \text{ or } d_4 = \sqrt{\frac{1}{2}\left(d_3^2 + r_2^2 + \frac{(d_3^2 + r_2^2)^2 - d_3^2 + r_2^2}{AB}\right)}$$

The first branch is the separating surface $d_4 = C_1$ between domains 1 and 2.

Equation (8) is a second-degree polynomial in $d_4$. Solving this quadratics for $d_4$ and assuming strictly positive values for $d_4$ and $r_2$ yields the following two branches for (8):

$$(d_4 = \frac{d_3}{d_3 - 1} \cdot B \text{ and } d_3 > 1) \text{ or } (d_4 = \frac{d_3}{1 - d_3} \cdot B \text{ and } d_3 < 1)$$

These two branches are the separating surfaces $d_4 = C_3$ and $d_4 = C_4$, respectively. In the same way, (9) can be rewritten as $d_4 = C_2$.

In conclusion, (5) and (6) do not define separating surfaces and only one branch of (7) defines a separating surface.

### 3.3 Number of nodes

In this section, we investigate each domain according to the number of nodes in the workspace.

*Domain 1:* Since all manipulators in this domain are binary, they cannot have any node in their workspace. Thus, all manipulators in domain 1 have the same workspace topology, namely, 0 node, 0 cusp and a hole inside their workspace. This workspace topology is referred to as $WT_1$ (Workspace Topology #1).

*Domain 2:* Figures 4 and 2 show two distinct workspace topologies of manipulators in domain 2, which feature 2 nodes and 0 node and which we call $WT_2$ and $WT_3$, respectively. Transition between these two workspace topologies is one such that the two lateral segments of the internal boundary meet tangentially (Fig. 9). Equation of this transition can be derived geometrically and the following equation is found (Baili, 2003):

$$d_4 = (A - B)/2 \tag{10}$$

As noted in section 3.2, a third topology exists in this domain, where the internal boundary exhibits a '2-tail fish'.

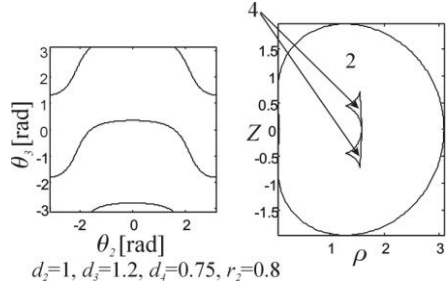 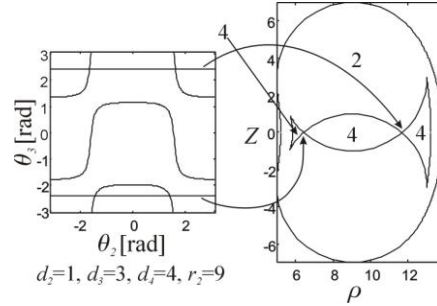

*Figure 9*. Transition $WT_2$-$WT_3$.    *Figure 10*. Workspace topology $WT_4$.

This workspace topology, which we call $WT_4$, features two nodes like in Fig. 4, but these nodes do not play the same role. They coincide with two isolated singular points, which are associated with the two singularity lines defined by $\theta_3=\pm\arccos(-d_3/d_4)$ (the operation point lies on the second joint axis and the inverse kinematics admits infinitely many solutions). Also, the nodes do not bound a hole like in Fig. 4 but a region with four IKS (Fig. 10).

Transition between $WT_3$ and $WT_4$ is a workspace topology such that the upper and lower segments of the internal boundary meet tangentially (Fig. 11). As shown in (Baili, 2003), this transition is the occurrence of the additional singularity $d_3 + c_3 d_4 = 0$. This transition is defined by:

$$d_4 = d_3 \qquad (11)$$

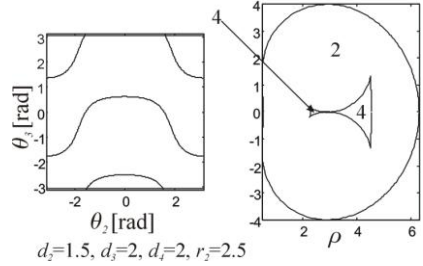

*Figure 11*. Transition $WT_3$-$WT_4$.

*Domains 3 and 5:* The internal boundary has either 2 cusps (domain 3) or 0 cusp (domain 5). This boundary is either fully inside the external boundary (like in Figs 5 and 7), or it can cross the external boundary, yielding two nodes like in Fig. 12. Thus, domain 3 (resp. domain 5) contains two distinct workspace topologies, which we call $WT_5$ (2 nodes) and $WT_6$ (resp. $WT_8$ and $WT_9$).

Transition between $WT_5$ and $WT_6$ and transition between $WT_8$ and $WT_9$ are such that the internal boundary meets the external boundary tangentially (Fig. 13). This transition can be derived geometrically and

the following equation is found (Baili, 2003):

$$d_4 = \frac{1}{2}(A+B) \qquad (12)$$

*Domains 4:* Manipulators in domain 4 have four cusps and four nodes. No subcase exists in this domain (Baili, 2003). Such topologies are referred to as $WT_7$.

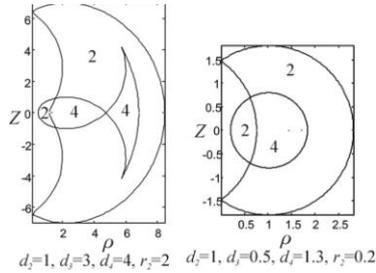 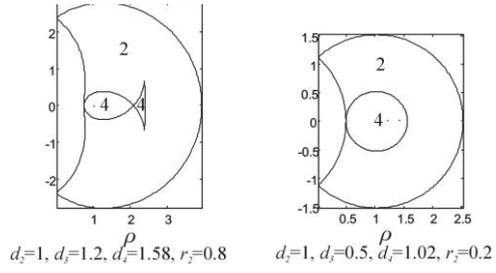

*Figure 12.* Workspace topologies $WT_6$ (left) and $WT_9$ (right)

*Figure 13.* Transition $WT_5$-$WT_6$ (left) and $WT_8$-$WT_9$ (right).

## 4. Results synthesis

The partition with cusps and nodes is shown in a section ($d_3$, $d_4$) of the parameter space for $r_2$=1(Fig. 14), where $E_1$, $E_2$ and $E_3$ are the right hand side of (10), (11) and (12), respectively. Plots of the separating curves in sections for different values of $r_2$ show that they deform smoothly with the same intersections when $r_2$ varies. The areas of $WT_1$, $WT_2$, $WT_7$ and $WT_9$ increase when $r_2$ decreases, whereas those of $WT_3$, $WT_4$, $WT_5$ and $WT_6$ decrease. The area of $WT_4$ is very narrow when $r_2$<1 and almost disappears when $r_2$<0.1.

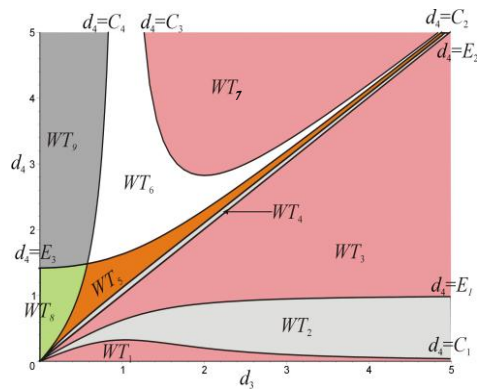 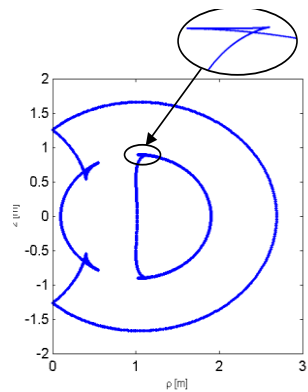

*Figure 14.* Parameter space partition in a section $r_2$=1.

*Figure 15.* A manipulator with 8 cusps when $r_3 \neq 0$.

## 5. Conclusions

A family of 3R manipulators was classified according to the topology of the workspace, which was defined as the number of cusps and nodes. The design parameters space was shown to be partitioned into nine subspaces of distinct workspace topologies. Each separating surface was given as an explicit expression in the DH-parameters. This study is being extended to manipulators with $r_3 \neq 0$. First results show that some may have 6 or 8 cusps (Fig. 15). But for small values of $r_3$, the partition is nearly the same as in Fig. 14. The subspace $WT_4$ does not exist any more. It is replaced by two adjacent tiny subspaces with 6 and 8 cusps. For high values of $r_3$, the partition gets complicated since more workspace topologies exist.